\documentclass[12pt]{article}
\usepackage{amsmath}
\usepackage{graphicx}
\usepackage{hyperref}
\usepackage[numbers]{natbib}
\usepackage{geometry}
\usepackage{booktabs,tabularx}
\title{First-place Solution for Streetscape Shop Sign Recognition Competition}
\author{Bin Wang, Li Jing 
\\ 
Zhejiang Gongshang University
\\
bingwang@mail.zjgsu.edu.cn
}

\date{December 2024}

\begin{document}

\maketitle

\begin{abstract}
Text recognition technology applied to street-view storefront signs is increasingly utilized across various practical domains, including map navigation, smart city planning analysis, and business value assessments in commercial districts. This technology holds significant research and commercial potential. Nevertheless, it faces numerous challenges. Street view images often contain signboards with complex designs and diverse text styles, complicating the text recognition process. A notable advancement in this field was introduced by our team in a recent competition. We developed a novel multistage approach that integrates multimodal feature fusion, extensive self-supervised training, and a Transformer-based large model. Furthermore, innovative techniques such as BoxDQN, which relies on reinforcement learning, and text rectification methods were employed, leading to impressive outcomes. Comprehensive experiments have validated the effectiveness of these methods, showcasing our potential to enhance text recognition capabilities in complex urban environments.

\end{abstract}
\section{Introduction}\label{Introduction}
\subsection{Background}\label{Background}
Storefront signboards ~\cite{tang2023character} are crucial for extracting essential information \cite{lu2024boundingboxworthtoken, zhang8, zhang9, zhang2024seeing, zhang2024simignore, zhang2024enhancing}  from street view imagery, making them significant research subjects in the realm of computer vision. The technology used for character recognition on these signboards integrates visual data processing with practical applications ~\cite{ic13,ic15,ic19mlt, shan2024mctbenchmultimodalcognitiontextrich, shen2024imagpose}, such as map navigation \cite{liu2022nommer}, smart city planning ~\cite{shen2024imagdressing,shenadvancing,shen2024boosting}, and commercial value analysis, presenting substantial research and commercial opportunities ~\cite{li2024enhancing, cao2023attention,cao2022gmn,liu2024hrvda}.
Participants in this competition are charged with the development of an advanced shop signboard recognition and analysis system \cite{Zhang01022023, 10.1145/3696271.3696294, 10.1145/3696271.3696292, 10.1145/3696271.3696299, 10692439, xiao-blanco-2022-people, xiao-etal-2023-context, xiao-etal-2024-analyzing, 10458651, 10.1007/978-3-031-63616-5_20, 10.1145/3655497.3655500,https://doi.org/10.1002/sam.11509}. Using street view photos and labels provided by the organizers, we face the challenge of precisely detecting shop signboards, analyzing their layouts, and recognizing shop names within the context of natural street view imagery. The competition is centered around natural street scenes \cite{bavcic2024towards, bavcic2024jy61, liu2024llm4gen, he2024mars, lu2024improving,liu2024omniclip,li2024frame,liu2024envisioning}, specifically targeting the storefront signboards and their textual content \cite{sun2025attentive, wang2025pargo}, with a focus on daytime images only. The given input is a street view image, and the required outputs are the locations of the shop signs (limited to visible signboards) and the names of the shops as depicted on these signboards. The primary objective of this competition is to significantly improve the accuracy of signboard detection \cite{liu2024cm, liu2024tolerant,liu2023lightweight,wang2024sparse,zhang2023body,jin2022imc} and the precision of the shop names recognized \cite{bi2024decoding, bian2024make, zhao2024metric, huo2024composition, yang2024llmcbench, liu2021context,liu2020jointly, liu2020saanet, liu2022memory, liu2021spatiotemporal, liu2022few, liu2024a, liu2024survey, liu2024pandora} in the test set.
\subsection{Challenge}
This competition presents several significant challenges:
\begin{itemize}
\item [1)]
\textit{Complex tasks.} The competition encompasses three intertwined tasks: signboard detection, text box detection, and text recognition.
\item [2)]
\textit{Complex scene.} The scenes to be analyzed are street views, which are inherently rich with diverse visual information. Moreover, a single image may contain multiple signboards amidst other non-signboard elements.
\end{itemize}
\subsection{Competition Details}
\subsubsection{Dataset}
\begin{figure}[t]
\centering
\includegraphics[width=1.0\linewidth]{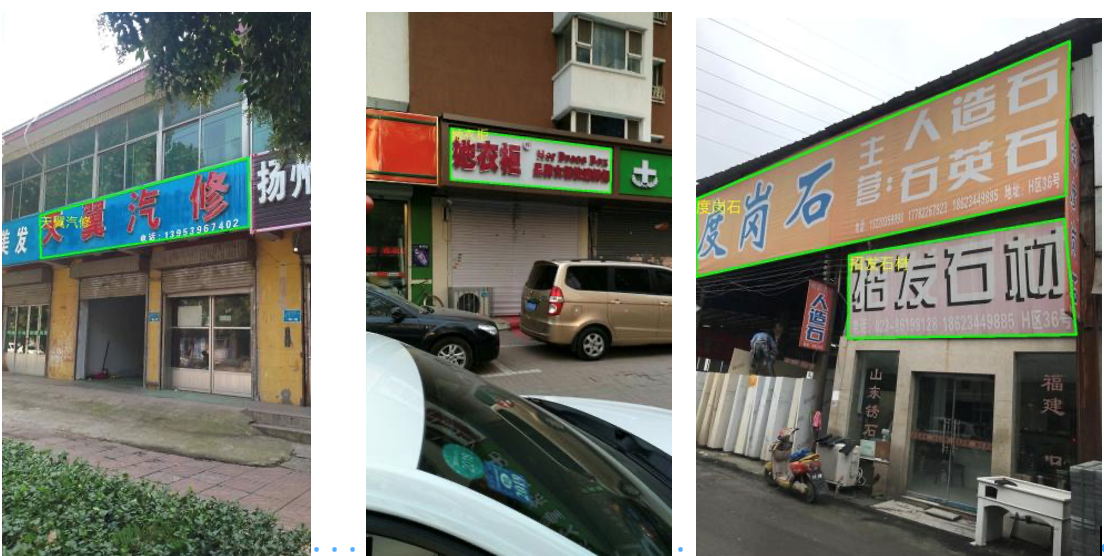}
\caption{Examples of store sign and store name data.}
\label{data1}
\end{figure}

\begin{figure}[t]
\centering
\includegraphics[width=1.0\linewidth]{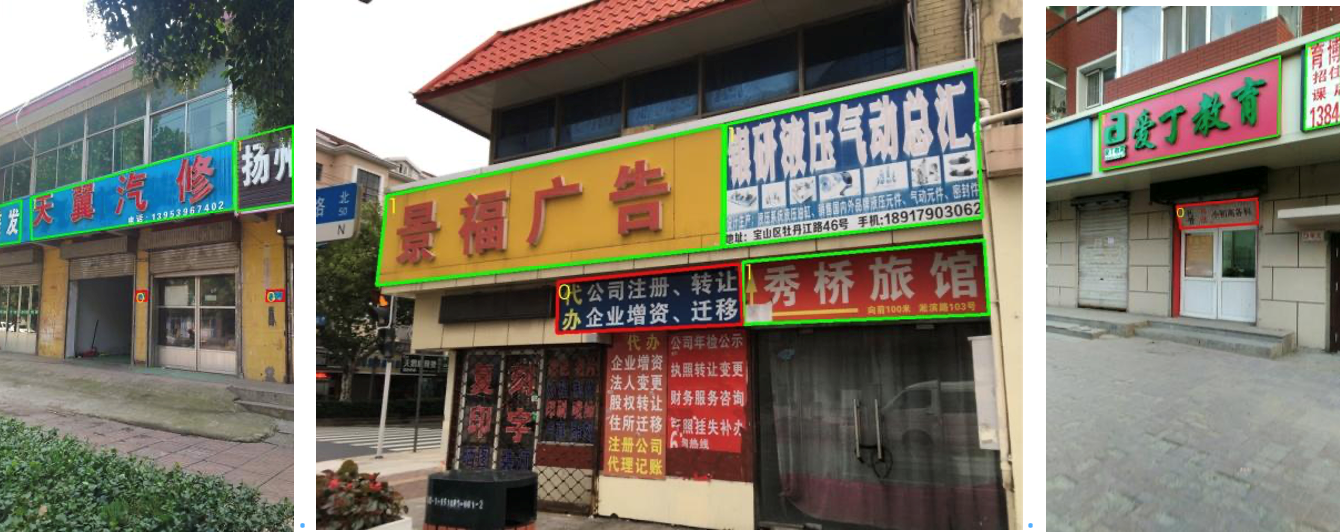}
\caption{Examples of signboard detection data.}
\label{data2}
\end{figure}

\begin{figure}[t]
\centering
\includegraphics[width=1.0\linewidth]{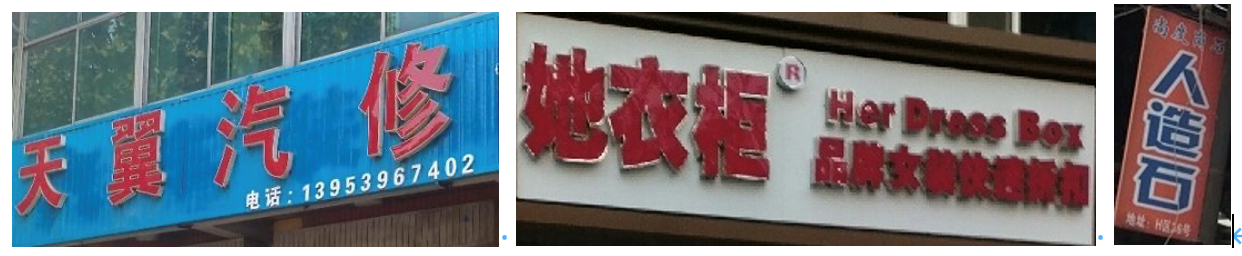}
\caption{Examples of signboard OCR data.}
\label{data3}
\end{figure}

Three distinct datasets have been provided for this challenge: "Store sign and store name data," "Signboard detection data," and "Signboard OCR data." The evaluation set will adhere to the format of the "Store sign and store name data." Below is an overview of each dataset:
\textbf{Store sign and store name data.} This dataset features street view images with potential multiple storefronts per image. Each storefront may display several signboards, some of which are unrelated to stores, and some that are explicitly store signboards. There is a possibility that some images do not contain any store signboards. Each image in the dataset is labeled with the valid quadrilateral box of the store signboard and its corresponding store name. Valid signboards are defined as having a complete and clear shape, minimal occlusion \cite{liu2021adaptive,liu2022unsupervised,liu2022imperceptible,liu2023point,liu2023hypotheses}, and displaying a specific store name. The training set comprises approximately 5,000 images, while the test set contains about 500 images. In the test set, store names are required to have at least two Chinese characters and must not include English characters or special symbols \cite{dan2024multiple, dan2024evaluation, dan2024image, yi2025score, he2024ddpm}. The label file is in text format, where each line specifies the coordinates of the four vertices of the signboard, arranged clockwise, and the corresponding store name. Sample images from the dataset are depicted in Fig. \ref{data1}.
\textbf{Signboard detection data.} Similar to the first dataset, this dataset includes street view images that may feature multiple storefronts and a variety of signboards, including non-store signboards. Some signboards may lack corresponding store names, as indicated in the first dataset, and some images may not feature any store signs at all. This dataset labels all signboards in each image with quadrangle boxes. The training data volume matches that of the first dataset. The label file format is text, with each line detailing the coordinates of the four vertices of the signboard frames, arranged clockwise, followed by a digit \cite{li2024real, mangin2024changing, li2024joint} indicating whether it is a non-store (0) or a store signboard (1). Samples from this dataset are shown in Fig. \ref{data2}.
\textbf{Signboard OCR data.} This dataset consists of close-up images of all signboards extracted from the aforementioned street view data. Each image may contain multiple lines of text. The labels specify the position and content of all text in each photo. This dataset encompasses 14,331 images. The label file is in TXT format, where each line details the coordinates of the four vertices of the text and the text content itself. Samples from this dataset are illustrated in Fig. \ref{data3}.

\section{Method}\label{Algorithm}


\begin{figure}[t]
    \begin{center}
    \includegraphics[width=1.0\linewidth] {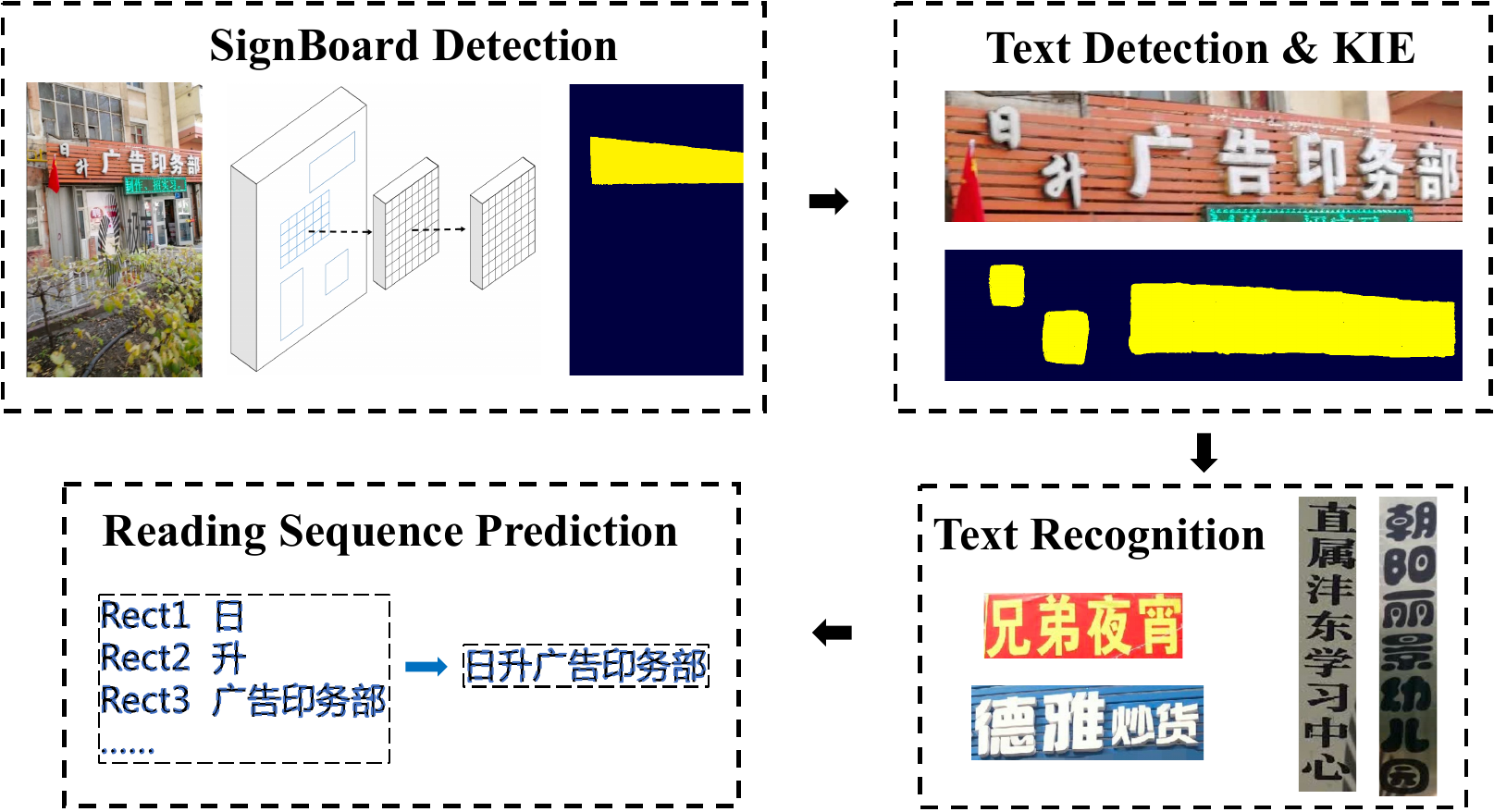}
    \end{center}
    \caption{The algorithm pipeline. KIE represents key information extraction. }
    \label{pipeline}
\end{figure}

 As shown in Fig.~\ref{pipeline}, our team uses a multi-stage algorithm, which is divided into four main stages: signboard detection, signboard text detection \& KIE, text recognition, and reading sequence prediction.

\subsection{Signboard Detection} 
During the signboard detection phase, the raw annotation data is meticulously reprocessed into three distinct categories: valid store signboards, invalid store signboards, and non-store signboards. This classification is essential because we need to identify the invalid shop signs among the signboards, even though the raw annotations typically only indicate whether they are store signboards. Given that the ground truth is represented as a quadrilateral region, we opted for instance segmentation \cite{2020ABCNet, tang2022few, fcenet} rather than traditional object detection \cite{faster-rcnn}. This approach allows us to segment the mask of the store signboard accurately. We then identify the smallest external quadrilateral encompassing the mask and apply a perspective transformation to this quadrilateral for text correction. The result is a precisely corrected signboard region.
The model architecture for instance segmentation, based on Mask-RCNN \cite{mask-rcnn}, has been enhanced in three key areas to boost its performance:
1) Integration of Deformable Convolutional Networks (DCN) \cite{DCN}: This addition aims to bolster the feature extraction capabilities of the model, allowing for more nuanced understanding and processing of the signboard images.
2) Keypoint Regression Branches Using Four Vertices: By treating the four vertices of the signboard's quadrilateral as keypoints, and adding keypoint regression branches, we enhance the model's accuracy and adaptability in a multi-task learning framework.
3) Advanced Data Augmentation Techniques: Beyond the standard augmentation methods, we implement 'copy-and-paste' and random perspective transformations to diversify the training data, thereby improving the model’s robustness and ability to generalize across varied real-world scenarios.
These strategic improvements are designed to refine the model's efficacy in detecting and analyzing store signboards accurately, making it a powerful tool in the realm of computer vision.

\begin{figure}[t]
    \begin{center}
    \includegraphics[width=1.0\linewidth] {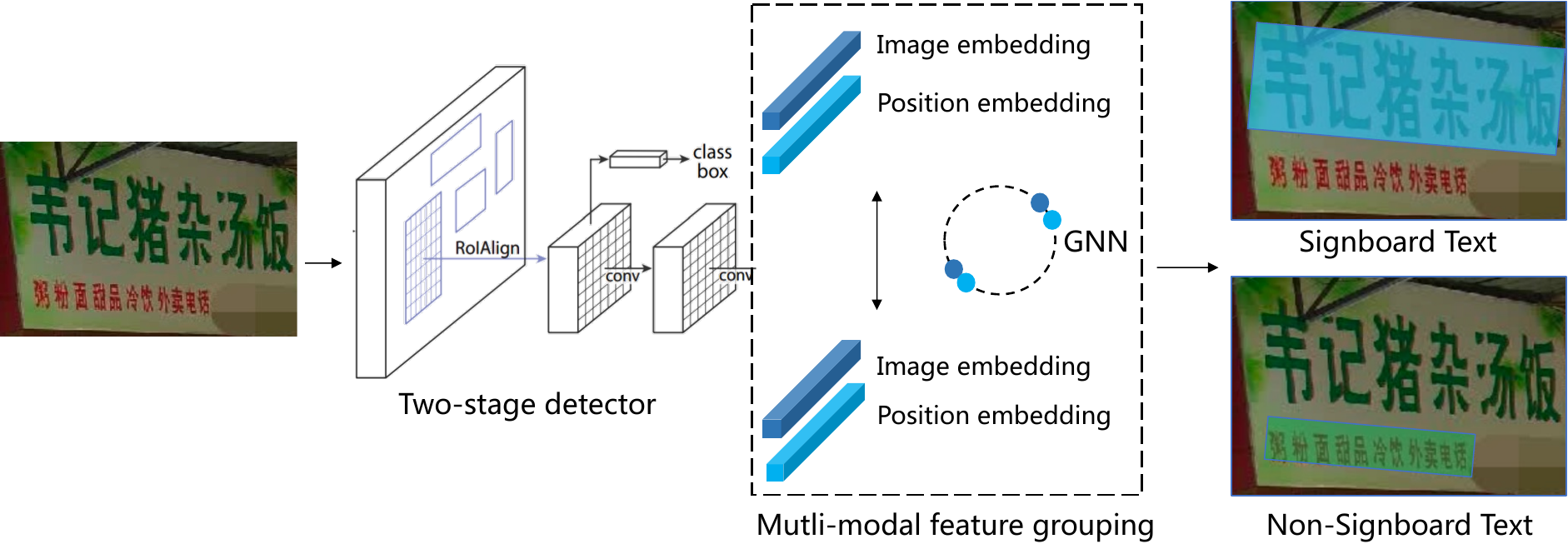}
    \end{center}
    \caption{The end-to-end network for joint text detection and KIE tasks. The model consists of two major parts. One is a two-stage detector that detects the position of the text and extracts the positional embedding and image embedding of the text. The other is a graph neural network for integrating multimodal features and distinguishing which text belongs to the store signboard.}
    \label{detection}
\end{figure}

\subsection{Signboard Text Detection \& Key Information Extraction (KIE)} In the stage of text detection~\cite{tang2022few, DB, PSE-Net, zhao2024harmonizing} and signboard text extraction phase, the two tasks are combined for end-to-end training due to the commonality of text features. The pipeline of the model is shown in Fig.~\ref{detection}. First, the model uses a two-stage detector to detect the location of the text and extract both the positional embedding and the image embedding of the text. A graph neural network is then used to integrate multimodal features and distinguish which belong to the store signboard texts. The end-to-end training paradigm, which is inspired by the former work~\cite{tang2022few}, allows for more powerful and effective features to be learned, thereby improving the accuracy of both tasks simultaneously. In addition, we use a reinforcement learning-based method called the BoxDQN~\cite{tang2022optimal}, which adjusts the shape of the text box to find the optimal box for the recognition model, thus improving the accuracy of end-to-end recognition.

\begin{figure}[t]
    \begin{center}
    \includegraphics[width=1.0\linewidth] {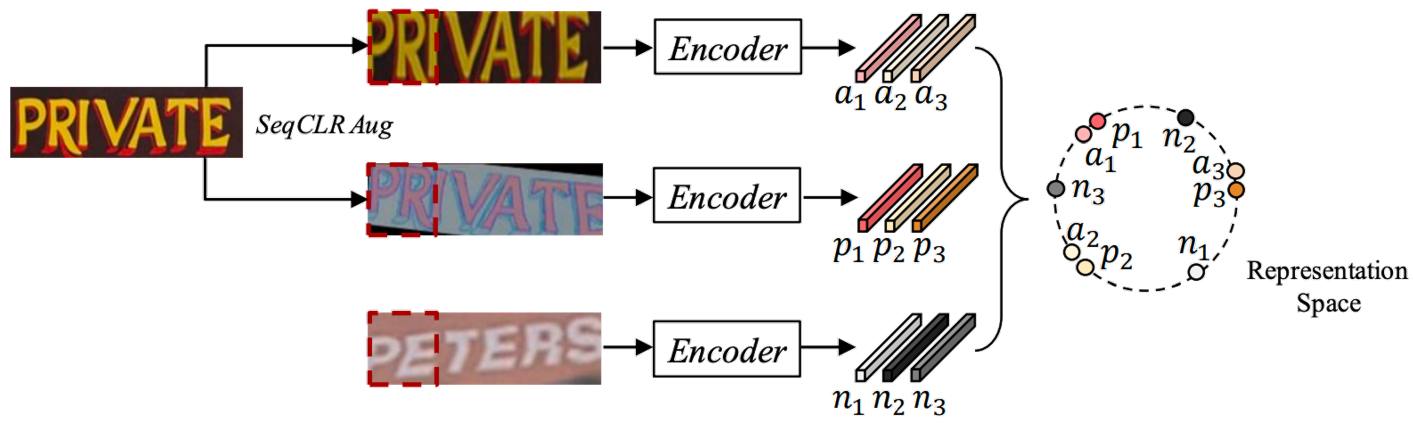}
    \end{center}
    \caption{Sequential Contrast Learning (SCL) for boosting scene text recognition. }
    \label{self-sup-1}
\end{figure}

\begin{figure}[t]
    \begin{center}
    \includegraphics[width=1.0\linewidth] {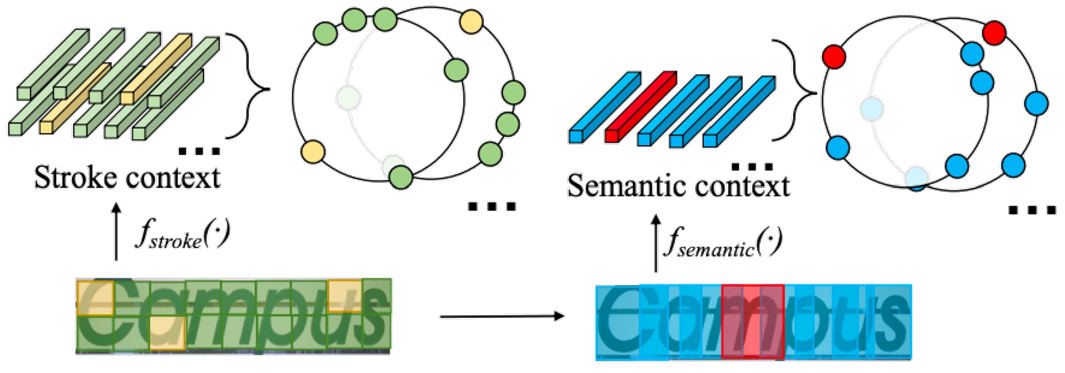}
    \end{center}
    \caption{Stoke and semantic context-based masked image modeling. }
    \label{self-sup-2}
\end{figure}

\begin{figure}[t]
    \begin{center}
    \includegraphics[width=1.0\linewidth] {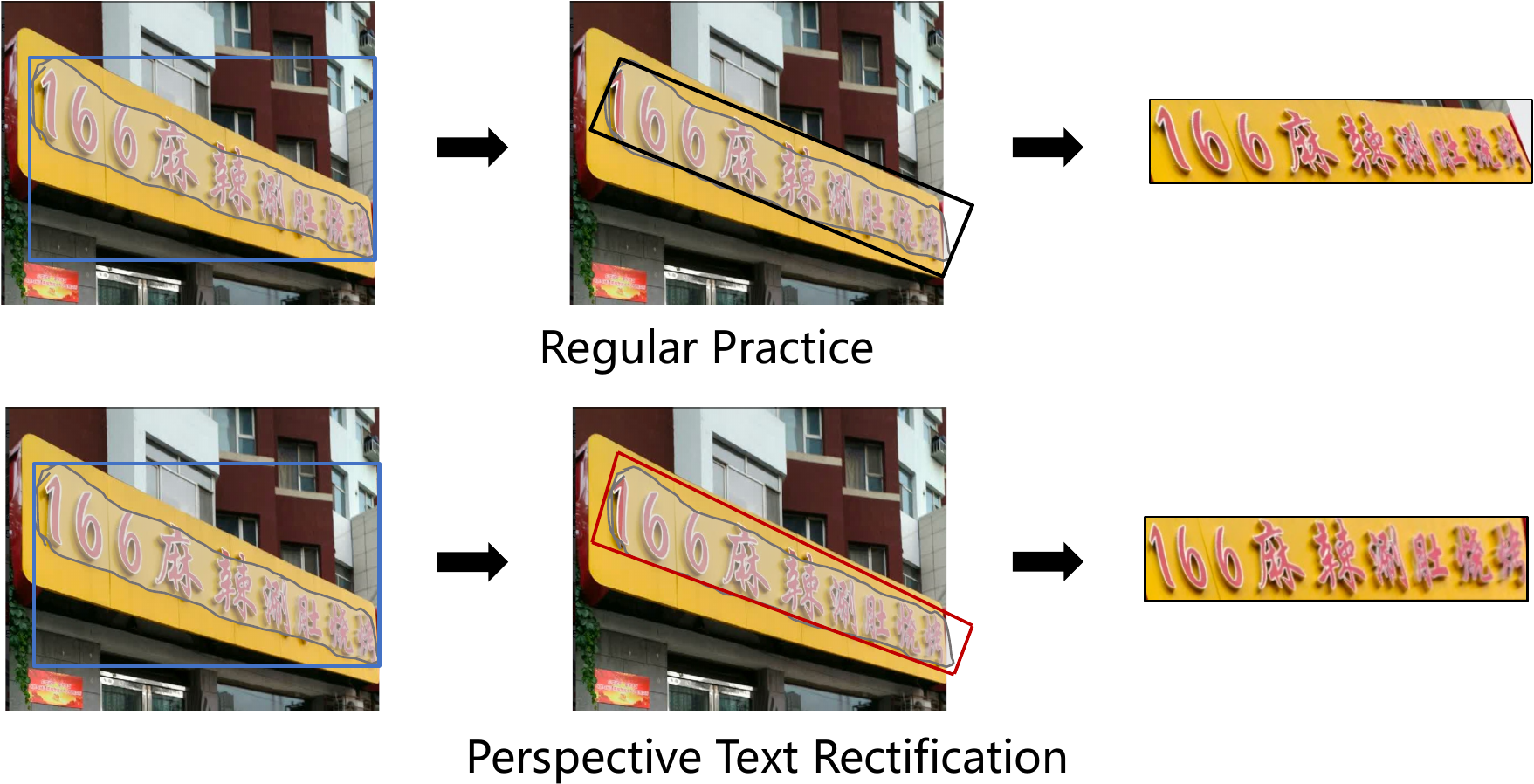}
    \end{center}
    \caption{Perspective text rectification. Compared with the conventional method of directly intercepting the minimum enclosing rectangle, our method is to obtain the minimum enclosing quadrilateral and then carry out perspective transformation to get a more accurate text region.}
    \label{per-text}
\end{figure}

\subsection{Text Recognition} In the text recognition phase, we judge the direction of the text based on the aspect ratio of the text box, and then train two text recognition models \cite{crnn, feng2023unidocuniversallargemultimodal, zhao2024multi, liu2023spts, tang2022youcan} for horizontal and vertical text, respectively. The baseline of the text recognition model we use is SAR~\cite{li2019show}. Based on it, we make some improvements. (1) upgrading the backbone to VIT~\cite{dosovitskiy2020image}, (2) changing the unidirectional LSTM to bidirectional LSTM \cite{lstm}, integrating LSTM and 2D attentional features in parallel, and providing richer contextual information. (3) using multiple self-supervised training, and (4) adding Center Loss to boost the discrimination of morphologically close word recognition features. Two self-supervised pre-training methods that contribute significantly are described below. The first one is a pre-training method based on Sequential Contrast Learning~\cite{aberdam2021sequence, sun2024attentiveeraserunleashingdiffusion}, as shown in Fig.~\ref{self-sup-1}. In detail, we perform some augmentation on the text image to get a positive sample pair, and then select another text image as a negative sample pair. The model extracts the sequence features using an encoder and then learns from the positive and negative sample pairs based on the sequence. The second one is masking pre-training based on stroke information and semantic context. As shown in Fig.~\ref{self-sup-2}, it is an improvement based on MAE~\cite{he2022masked}, where different masks are set for stroke features and semantic features to predict and learn the masks. In this way, the mural information and textual information of the text can be modeled simultaneously to improve the feature representation of the model. Using these two different self-supervised pre-training methods, we can obtain an extractor with strong modeling ability, thus improving the performance of text recognition.

\subsection{Reading Sequence Prediction} In the reading sequence prediction stage, since the store signboard text does not always consist of only one text block, it needs to be merged according to the relative position of the text blocks. We simply follow the logical relationships for text box reading sequence prediction without any use of a neural network.

\subsection{Results} 
Overall, our approaches can be found in Tab.~\ref{tab_method_performance}. In this competition, we incorporate three techniques that have become popular in recent years: multimodal feature fusion, large-scale self-supervised training, and a Transformer-based large model. In addition, depending on the properties of the data provided, we add some illuminating methods such as BoxDQN based on reinforcement learning and text rectification, which achieve promising performance.

\begin{table}[h]
		\centering
		\setlength\tabcolsep{10pt}
		\caption{The methods that are applied in the competition and our contribution.}
		\begin{tabularx}{1.0\linewidth}{c|c}
			\hline
			Method & Gain (\%)\\
			\hline
			DCN ~\cite{DCN} & 0.6\\
			BoxDQN~\cite{tang2022optimal} & 1.6 \\
			VIT~\cite{dosovitskiy2020image} backbone & 2.5\\
			Self-supervised pre-training & 3.2\\
			Mutli-modal modelling \& GNN & 2.9\\
			Perspective text rectification & 2.6 \\
			Center Loss for text recognition & 0.8\\
			\hline
		\end{tabularx}	
		\label{tab_method_performance}
\end{table}

\section{Experiments}\label{Reason}

\subsection{Evaluation Metric}

In order to support domestic machine learning development, the competition uses the Jittor\cite{hu2020jittor} framework to provide the baseline code of the competition.
Jittor is a high-performance deep learning framework based on instant compilation and meta operators. The entire framework integrates powerful Op compilers and tuners with instant compilation to generate customized high-performance code for your model. Jittor also contains a rich, high-performance model library, covering image recognition, detection, segmentation, generation, differentiable rendering, geometry learning, reinforcement learning, etc.

The evaluation metrics of the competition are the accuracy and recall rates of store signboard identification. Only when the location of the detection frame of the shop signboard is accurate and the identification of the shop name line in the signboard is correct, is the identification defined to be accurate. The formula for calculating the recall rate of store sign recognition is the ratio of the number of correctly recognized store signs to the true number of store signs. The calculation formula of the accuracy rate of store sign recognition is the ratio of the number of correctly recognized store signs to the total number of recognized store signs. The final score is the F-score of accuracy rate and recall rate, calculated as $2 * accuracy * recall/(accuracy + recall)$.

\subsection{Competition Results}
There are many challenges in street view store signboard recognition tasks, such as various modes and contents of store signboards, geometric distortion and truncation in image shooting, and limited training samples. Our team, with its profound accumulation in the OCR field for many years, has better analyzed the difficulties and proposed a reasonable algorithm framework according to the characteristics of the competition, and won first place in the competition. The overall framework divides the competition task into four stages: signboard detection, signboard text detection, text recognition, and reading sequence prediction. The main highlights are" signboard text detection and key information extraction", "multi-modal modeling \&GNN "," text self-supervised pre-training", and "VIT model". This method effectively integrates multi-dimensional information such as image features and text position distribution for multi-modal recognition and reasoning. In the case of limited samples, self-supervised pre-training and data augmentation methods are used to improve the performance significantly in a short period of time.

Finally, our team won first place among 14 teams with an F-score of 0.6672, an oral defense score of 92.86, and a combined score of 97.11.
With a reasonable design, this framework can be quickly implemented and applied in the industry and can play a good role in restoring the real world and building more realistic maps.

\section{Conclusion}\label{Future}

The profound development of deep learning has catalyzed significant advancements in the field of computer vision. Previously confined to elementary tasks such as detection, segmentation, and classification, the understanding of images has now expanded to encompass scene understanding, multi-modal recognition, relational reasoning, and comprehensive end-to-end image interpretation.
In the specific context of text recognition in street view store signboards, computer vision research has identified several key evolving trends:
1) Adoption of Semi-supervised or Unsupervised Learning: Large-scale pre-training techniques that rely on semi-supervised or unsupervised learning have become prevalent. These methods focus on deciphering the subtle relationships and intrinsic connections within roadway data, providing robust pre-training representations that bolster downstream perception tasks. This approach not only reduces the necessity for finely labeled data but also enhances the model's generalization capabilities and diminishes the costs associated with research and development.
2) Multi-dimensional, Multi-modal Feature Analysis: The use of sophisticated multi-modal, multi-task object relationship analysis based on multi-dimensional features has proven effective in elevating recognition accuracy.
3) Enhanced Scene and Quality Recognition: Advanced scene recognition and quality assessments are utilized to sift through and exclude non-relevant data, ensuring that only pertinent information is processed.
4) Progress in End-to-End Image Understanding: Rapid advancements in end-to-end methodologies are setting the stage for the emergence of real-world map auto-generation technologies. These sophisticated systems integrate targeted content analysis, detailed scene evaluation, and precise location restoration.
Looking ahead, these innovations are poised to significantly impact various sectors, playing a pivotal role in navigation systems, intelligent recommendations, smart city development, and autonomous driving technologies. In the near future, with the development of multimodal large models ~\cite{bai2023qwen, tang2024textsquare, zhao2024tabpedia, feng2025docpedia, wang2025pargo}, end-to-end signature text reading becomes possible with a higher ceiling.

\bibliographystyle{plainnat}
\bibliography{main}
\end{document}